%% file: main.tex
\pdfoutput=1

\documentclass{article}
\usepackage{colm2024_conference}   %

\usepackage[utf8]{inputenc}
\usepackage[T1]{fontenc}
\usepackage{url}
\usepackage{booktabs}
\usepackage{comment}
\usepackage{amsfonts}
\usepackage{nicefrac}
\usepackage{microtype}
\usepackage{xcolor}

\usepackage{subcaption}
\usepackage{makecell}
\usepackage{amsmath}
\usepackage{amssymb}
\usepackage{amsthm}
\usepackage{colortbl}
\usepackage{empheq}
\usepackage[many]{tcolorbox}
\usepackage{stfloats}
\usepackage{tabularx}
\usepackage{multirow}
\usepackage{bbm}
\usepackage{pifont}
\usepackage{array}
\usepackage{wrapfig}

\definecolor{pblind}{HTML}{5E7D95}
\definecolor{psat}{HTML}{A0803F}
\definecolor{pspur}{HTML}{9E565A}
\definecolor{pagree}{HTML}{7A8B7F}
\definecolor{ink}{HTML}{171522}
\definecolor{muted}{HTML}{625E70}
\definecolor{rulegrey}{HTML}{E2E1E6}
\definecolor{panelgrey}{HTML}{F7F7F9}
\definecolor{brandv}{HTML}{615CED}
\definecolor{memhl}{HTML}{1C6B5E}

\usepackage{tikz}
\usetikzlibrary{arrows.meta,positioning,calc,fit,backgrounds,shapes.geometric,decorations.pathreplacing,patterns}

\usepackage{hyperref}       %
\usepackage{cleveref}       %

\usepackage{silence}
\input{commands.tex}

\newcommand{\ourmethod}{\texttt{HarnessRL}}
\newcommand{\ourmethodbf}{\textbf{\texttt{HarnessRL}}}
\newcommand{\harness}{\ensuremath{\mathcal{H}}}
\newcommand{\corpus}{\texttt{Rubric-Mix-10k}}

\newcommand{\policyt}{\ensuremath{\pi_{\theta_t}}}

\usepackage{calc}
\newcommand{\best}[1]{\textbf{#1}}
\newcommand{\snd}[1]{\underline{#1}}
\colorlet{oursfg}{mypurple}
\colorlet{oursbg}{oursfg!8}
\definecolor{gaingreen}{RGB}{34,139,34}
\newcommand{\gain}[2]{\textbf{#1}{\fontsize{5}{5}\selectfont\color{gaingreen}(+#2)}}
\newcommand{\tabhead}[1]{#1}
\newcommand{\tabgroup}[1]{\textit{#1}}
\newcommand{\tabtop}{\toprule}
\newcommand{\tabbottom}{\bottomrule}
\newcolumntype{Y}{>{\centering\arraybackslash}X}
\newcommand{\tabsrc}[2]{\begin{tabular}[t]{@{}l@{}}#1\\[-1pt]{\scriptsize #2}\end{tabular}}
\newcommand{\tabsub}[2]{\begin{tabular}[t]{@{}l@{}}#1\\[-1pt]{\scriptsize #2}\end{tabular}}
\title{\raisebox{-2.4pt}{\includegraphics[height=13pt]{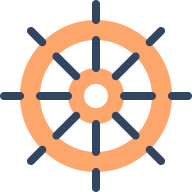}}\,\ourmethodbf{}: Co-Evolving Policies with a Just-in-Time Training Harness for Open-Ended RL}

\hypersetup{breaklinks,colorlinks=true,linkcolor=dblue,citecolor=dblue,urlcolor=dblue}
\newcommand*\thankstarget{\protect\hypertarget{Hfootnote.\number\value{footnote}}{}}
\newcommand*\samethanks[1][\value{footnote}]{\hyperlink{Hfootnote.\number#1}{\footnotemark[#1]}}

\author{%
Xiaoqiang Wang\thanks{\thankstarget Work done during an internship at Alibaba Token Hub, Alibaba Group.},
Ting-En Lin,
Yuchuan Wu,
Pinyi Zhang,
Caixin Kang,
Ling Li,\\
Bingxue Qiu,
Jieping Ye,
Yongbin Li\thanks{\thankstarget Corresponding authors.},
Bang Liu\samethanks{}
\\[0.6em]
{\large Alibaba Token Hub, Alibaba Group}\\[0.2em]
{\large Universit\'e de Montr\'eal \quad Mila - Quebec AI Institute}
}

\begin{document}

\maketitle

\begin{abstract}
Rubric-based reinforcement learning (RL) trains policies for open-ended generation with natural-language criteria as rewards, but these rewards can become ineffective as training progresses. We identify two failure modes. Under \emph{non-discriminative rewards}, scores collapse to uniformly low or high values due to insufficient exploration or a mismatch between task difficulty and policy capability. Under \emph{spurious rewards}, higher scores do not correspond to better responses.
Both reflect a training configuration whose task sampling, exploration, and evaluation stay fixed as the policy evolves.
We introduce \textbf{\ourmethod{}}, a \emph{training-time harness} that co-evolves with the policy through three interfaces for task sampling, rollout guidance, and rubric criteria. \ourmethod{} maintains a training-time memory from which a harness evolver retrieves within-task history and cross-task evidence. Using this evidence, the evolver attributes each reward failure to the interface whose update can address it. It then runs a proposer and just-in-time critic loop in which the critic accepts a candidate update only after measuring its effect on the current frozen policy, returning rejected candidates for revision.
We also curate \corpus{}, a five-domain training and evaluation suite for open-ended RL across medical, science, writing, role-play, and instruction-following tasks. On \corpus{}, \ourmethod{} outperforms the strongest baselines by 2.8 to 3.1 points across both 4B and 9B model scales and nearly halves the training steps to its best checkpoint.\footnotemark
\end{abstract}
\footnotetext{Code and data are available at \url{https://github.com/Tongyi-ConvAI/Qwen-Character/tree/main/HarnessRL}.}

\section{Introduction}
\label{sec:intro}

Reinforcement learning with verifiable rewards (RLVR)~\citep{lightman2023verify,lambert2024tulu,guo2025deepseekr1_cased} has become a standard recipe for LLM post-training~\citep{ouyang2022training_fix,bai2022training_fix,grattafiori2024llama3_fix,yang2025qwen3,deepseekai2026deepseekv4_fix}, improving mathematical reasoning~\citep{shao2024deepseekmath_cased,jaech2024openai_fix,wang2026system}, code generation~\citep{pan2024swegym,jain2024livecodebench_fix,deng2025swe_fix}, and agentic tasks~\citep{wang2025oscar,jin2025searchr1,liu2025mlagent,yang2026frontisma1,falck2026replicate}.
However, open-ended role-playing~\citep{wang-etal-2024-rolellm,wang2025coser,kang2026perception}, long-form writing~\citep{wu2025writingbench,bai2025longwriter}, and social reasoning~\citep{zhou2024sotopia,kang2025can,gong2026vinci2} lack rule-based verifiers because no answer is uniquely correct and quality is multidimensional~\citep{wang-etal-2024-fac2e,shan2026rubricrlsurvey} and often subjective. Rubrics as rewards~\citep{gunjal2025rar,huang2025rubicon,viswanathan2025rlcf,liu2025openrubrics} supply a reward for these tasks by decomposing quality into LLM-judged natural-language criteria~\citep{li2026rubrichub,chen2026rubricsurvey}.

\begin{figure}[t]
\centering
\begin{minipage}[c]{0.466\textwidth}
  \centering
  \includegraphics[width=\linewidth]{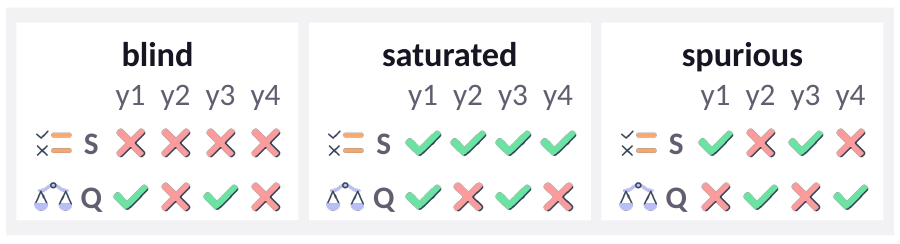}\\[1.1mm]
  \includegraphics[width=\linewidth]{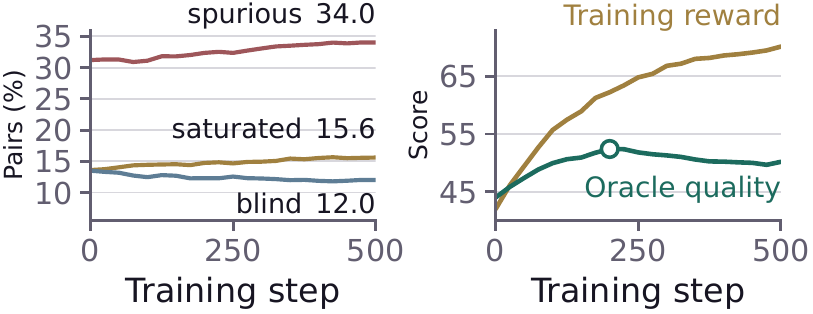}
\end{minipage}\hfill
\begin{minipage}[c]{0.504\textwidth}
  \centering
  \includegraphics[width=\linewidth]{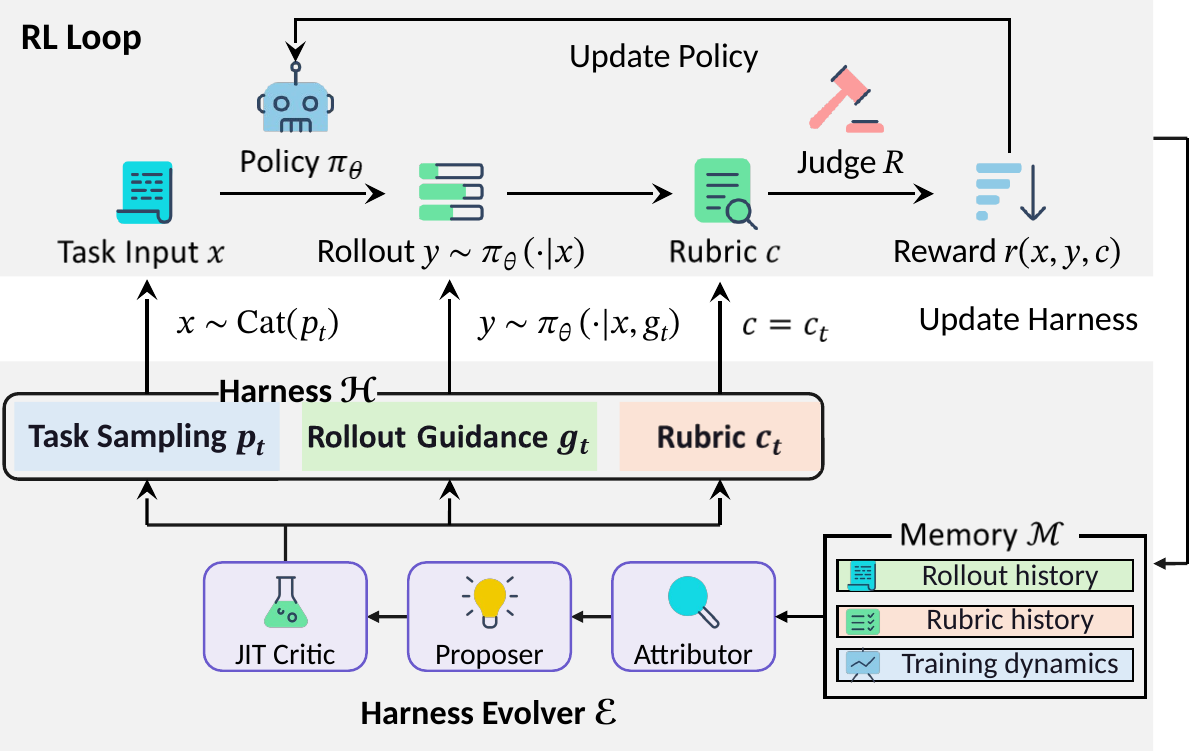}
\end{minipage}
\caption{%
Rubric-reward failures under GRPO on \corpus{} at \mbox{Qwen3.5-4B}, and harness-controlled RL training in \ourmethod{}: failure classes for one criterion (top left), class proportions and GRPO's training reward against oracle quality on held-out tasks (bottom left), and harness $(p_t,g_t,c_t)$-controlled RL training (right).
$S$ denotes the judge's binary criterion scores and $Q$ the oracle ranking of the rollouts.}
\label{fig:motivation}
\vspace{-5mm}
\end{figure}

However, rubric rewards can become unreliable as training progresses. Policies may exploit superficial cues, including opening praise and a closing claim that every requirement is satisfied~\citep{huang2025rubicon,wang2026rubrichacking}, raising rubric scores while quality stalls. Fixed rubrics may not distinguish useful behaviors, while on-policy rollouts may miss rewarded ones. Recent methods evolve rubrics~\citep{guan2026evorubric,wu2026amaris,wu2026dynamicrubric} or use privileged rollout guidance~\citep{zhou2025ruscarl,xia2026skillrl_fix,huang2026ocgrpo} for exploration. Each adapts the rubric or the rollout guidance without diagnosing the reward failure, yet adapting either one alone can shift the failure to another part of training. Guidance that helps the policy meet a fixed rubric can saturate its criteria, and a saturated group carries no advantage~\citep{yu2026dapo}. Conversely, a rubric tightened without guidance can place criteria beyond the policy's reach, where they give no learning signal~\citep{sundaram2026learnability}.

Rubric-reward failures fall into two modes. \textbf{Non-discriminative} rewards collapse to uniformly low or high scores, \ie \clblind{} or \clsat{}, leaving little relative learning signal within a rollout group. \textbf{Discriminative but spurious} rewards, \ie \clspur{}, continue to separate responses, but their induced preferences do not align with response quality. We measure quality by the oracle ranking of the oracle judge, a stronger LLM judge that compares responses pairwise within each group under one quality guideline per domain (\Cref{sec:exp}). These failures account for $61.6\%$ of response pairs in a rubric-based GRPO~\citep{shao2024deepseekmath_cased} run (\Cref{fig:motivation}). Of all pairs, $27.6\%$ provide no usable contrast, while $34.0\%$ provide contrast that does not reliably track quality, including $19.8\%$ that the reward ranks in the wrong direction. Directly optimizing the observed rubric reward can thus provide an ineffective or even misleading learning signal.

The three failure classes call for updates to different parts of training, rollout guidance for a \clblind{} criterion, task sampling for a \clsat{} task, and the rubric for a \clspur{} criterion. We therefore introduce \ourmethod{}, which treats task sampling, rollout guidance, and rubric criteria as three interfaces of a \textbf{training-time harness \harness{}} that co-evolves with the policy. \ourmethod{} reallocates sampling toward tasks with useful learning signal, injects rollout guidance when exploration is insufficient and withdraws it as training progresses, and evolves rubric criteria to correct coarse, saturated, or spurious rewards.

\harness{} uses a training-time memory $\mathcal{M}$ that records rollouts, the sampling, rollout and rubric updates, and their outcomes. The harness evolver $\mathcal{E}$ retrieves within-task history from this memory to detect recurring failures and track the effects of earlier updates, and cross-task evidence to transfer examples between tasks. Using this evidence, it attributes each reward failure to the interface whose update can address it, then runs a proposer and just-in-time (JIT) critic loop. The critic accepts a candidate update only after measuring its effect on the current frozen policy, and accepts a rollout or rubric update only when that effect lies within the corresponding calibrated acceptance band. Rejected updates return with their measurements to the proposer, which proposes \texttt{noop} when no update is supported.

We also curate \corpus{} for training and held-out in-domain and out-of-distribution (OOD) evaluation across medical, science, writing, role-play, and instruction-following tasks. On \corpus{}, \ourmethod{}-trained Qwen3.5-4B~\citep{qwen35_2026} scores 64.5 in domain and 49.8 OOD, exceeding the strongest baseline by 3.0 and 2.8 points. With Qwen3.5-9B, the margins are 2.9 points in domain and 3.1 points OOD. At the end of training, reward-oracle agreement is 48.9\% under \ourmethod{} against 38.4\% under GRPO. \ourmethod{} also reaches its best OOD checkpoint before every adaptive-training baseline.

\begin{figure}[t]
\centering
\includegraphics[width=\textwidth]{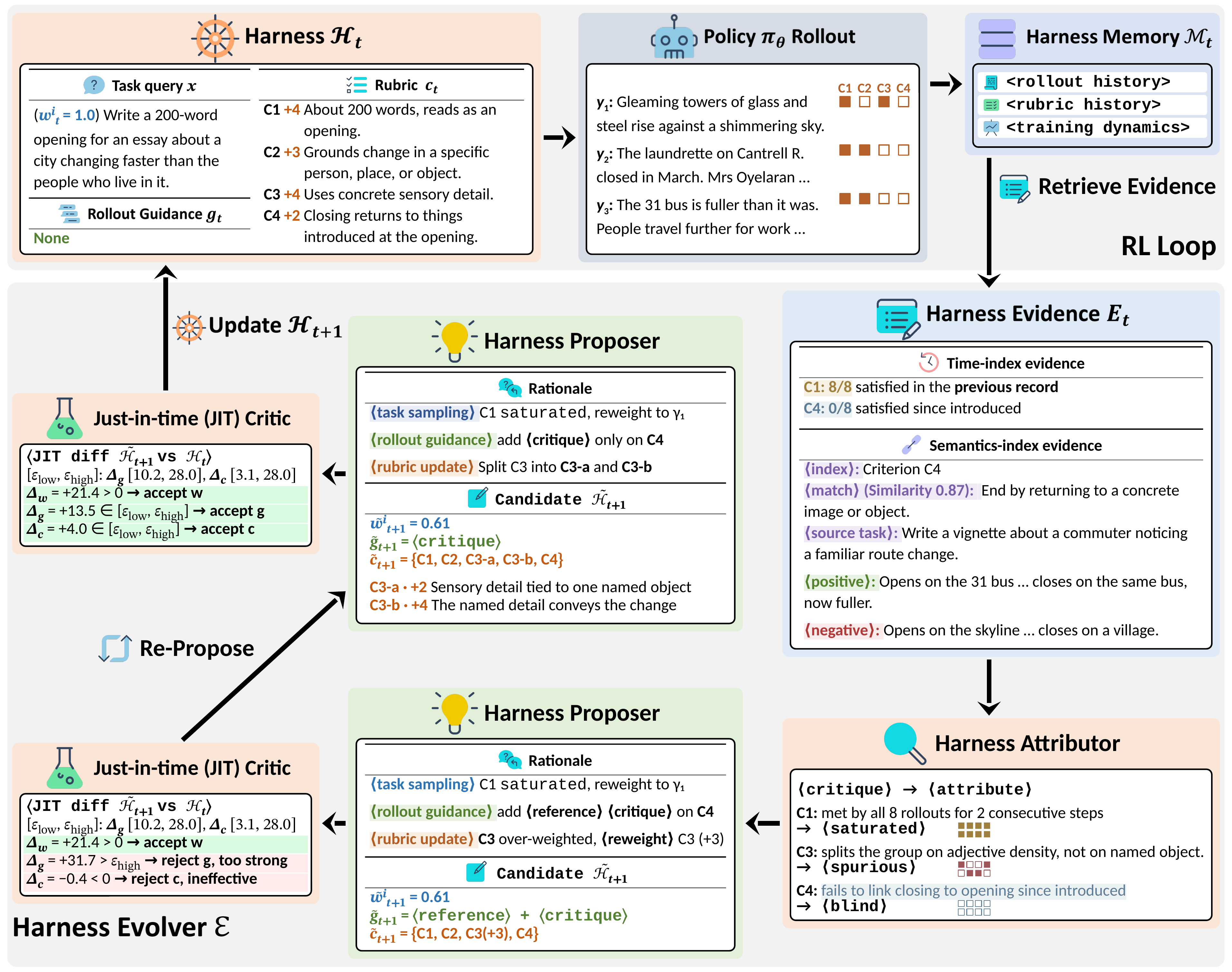}
\caption{%
\ourmethod{} runs an RL loop conditioned on a training-time harness $\harness{}_t=(p_t,g_t,c_t)$, which controls task sampling, rollout guidance and the rubric, respectively.
The harness co-evolves with the policy: it retrieves evidence from the training-time memory $\mathcal{M}_t$, attributes failures, and is iteratively refined by a proposer and JIT critic loop.}
\label{fig:pipeline}
\vspace{-5mm}
\end{figure}

\section{\ourmethod{}}
\label{sec:method}

\ourmethod{} augments a standard RL loop with an adaptive \emph{training-time harness} $\mathcal{H}_t=(p_t,g_t,c_t)$ that controls three interfaces (\Cref{fig:pipeline}). The sampling interface $p_t$ distributes probability over a fixed task bank $\mathcal{D}$ toward tasks with useful learning signal. The rollout interface $g_t$ injects privileged guidance when the policy cannot reach rewarded responses and withdraws it as the policy improves. The rubric interface $c_t$ specifies the criteria under which a fixed judge scores each response.

As the policy learns, \ourmethod{} adapts $\harness{}$ in two steps. \emph{Failure attribution} assigns each reward failure a class and routes it to the interface whose update can address it. \emph{Measured intervention} accepts a candidate update only after measuring its effect on the current frozen policy. The \textbf{harness memory} $\mathcal{M}$ supplies evidence for both steps. In the \textbf{harness evolver} $\mathcal{E}$, the attributor assigns labels, the proposer constructs candidate updates, and the JIT critic measures and accepts or rejects them. The accepted updates form $\harness{}_{t+1}$.

\subsection{Harness-Controlled RL Training}
\label{sec:harness_training}

At step $t$, the harness $\mathcal{H}_t=(\hctrl{p_t},\hctrl{g_t},\hctrl{c_t})$ conditions the inner RL loop through task sampling, rollout guidance, and rubric criteria. Let $\mathcal{D}=\{x_i\}_{i=1}^{N}$ be the training task bank and $\policyt$ the current policy. Each step draws $512$ tasks from $\hctrl{p_t}$. Because the harness evolver analyzes each draw independently, we trace index $i$ and write $x_t=x_i$. The policy samples $Y_t=\{y_t^{(j)}\}_{j=1}^{J}$, and the fixed judge $R$ scores each rollout against the current rubric's $K_t$ criteria $q_{t,m}$,
\begin{equation}
    x_t \sim \operatorname{Cat}(\hctrl{p_t}), \qquad
    y_t^{(j)} \sim \policyt(\cdot \mid x_t,\hctrl{g_t}), \qquad
    s_{t,m}^{(j)} = R\bigl(x_t,y_t^{(j)},q_{t,m}\bigr),
    \quad j=1,\ldots,J .
\end{equation}

\noindent\textbf{Sampling interface.} Tasks the policy has saturated provide little learning signal, so the sampling interface reallocates probability toward tasks that remain useful~\citep{chen2025sec,zeng2026cures}. We parameterize $\hctrl{p_t}$ by non-negative task weights, $p_t^i=w_t^i/\sum_{\ell=1}^{N}w_t^\ell$, so $\operatorname{Cat}(\hctrl{p_t})$ selects $x_i$ with probability $p_t^i$. The harness evolver sets $w_t^i$ from that task's training dynamics. Because reallocation alone would change the objective, each draw carries $u_t=1/(Np_t^i)$, the importance-sampling correction of prioritized experience replay~\citep{schaul2016prioritized}. This preserves each task's expected contribution under the original uniform distribution, so reweighting changes only its sampling frequency.

\noindent\textbf{Rollout interface.} Positive reward is rare while the policy cannot reach rewarded responses. Building on scaffolded RL exploration~\citep{yu2026dapo,zhou2025ruscarl}, the rollout interface steers exploration with privileged guidance. For each task, $\hctrl{g_t}\subseteq\{f_1,f_2,f_3\}=\{\gtag{criterion},\gtag{reference},\gtag{critique}\}$ contains the guidance fields mounted after $x_t$ in index order. The $\gtag{criterion}$ field states a constraint paraphrased from the target criterion or, when available, taken from earlier behavior that satisfied it. The $\gtag{reference}$ field holds positive and negative responses from the task history or a donor task with a matching criterion. The $\gtag{critique}$ field records the previous group's shortfall. The seed harness $\harness{}_0$ mounts no fields. The evolver selects each task's fields and withdraws them as the policy internalizes the demonstrated behavior. A schedule removes remaining fields by training's end, so $\hctrl{g_t}=\emptyset$ at evaluation.

\noindent\textbf{Rubric interface.} A fixed rubric can reward superficial cues unrelated to response quality, producing \clspur{} criteria. The rubric interface $\hctrl{c_t}=\{(q_{t,m},\lambda_{t,m})\}_{m=1}^{K_t}$ is a weighted set of natural-language criteria, where $q_{t,m}$ is the $m$-th criterion and $\lambda_{t,m}\neq0$ its signed weight. The judge $R$ evaluates each criterion independently, producing binary scores $s_{t,m}^{(j)}$ in the $K_t\times J$ matrix $S_t$ read by the harness attributor. The weights give reward $r_t^{(j)}=\sum_{m}\lambda_{t,m}\,s_{t,m}^{(j)}\,/\sum_{m}\max(\lambda_{t,m},0)$. Negative weights penalize undesirable behavior, while normalization by total positive weight preserves a common scale across rubric sizes. For an attributor-labeled \clspur{} criterion, the harness evolver uses the rubric operations in \Cref{sec:evolver} to rewrite $q_{t,m}$, reweight $\lambda_{t,m}$, or create, delete, split, and merge criteria.

\noindent\textbf{RL objective.} We extend GRPO~\citep{shao2024deepseekmath_cased} to draw tasks from $\hctrl{p_t}$, generate rollouts under $\hctrl{g_t}$, compute rewards under $\hctrl{c_t}$, and weight each task loss by $u_t$,
\begin{equation}
\begin{aligned}
\mathcal{J}(\theta_t)
&= \mathbb{E}_{x_t\sim\operatorname{Cat}(\hctrl{p_t}),\;Y_t\sim\pi_{\theta_{\mathrm{old}}}(\cdot\mid x_t,\hctrl{g_t})}
   \Bigl[u_t\,\mathcal{L}\bigl(\theta_t;x_t,Y_t,\hctrl{c_t}\bigr)\Bigr]
   -\beta\,\mathbb{D}_{\mathrm{KL}}\bigl(\policyt\,\|\,\pi_{\mathrm{ref}}\bigr),\\
\mathcal{L}\bigl(\theta_t;x_t,Y_t,\hctrl{c_t}\bigr)
&= \frac{1}{J}\sum_{j=1}^{J}\frac{1}{|y_t^{(j)}|}\sum_{k=1}^{|y_t^{(j)}|}
   \min\bigl(\rho_{j,k}\hat{A}_t^{(j)},\,
   \mathrm{clip}(\rho_{j,k},1{-}\epsilon,1{+}\epsilon)\hat{A}_t^{(j)}\bigr),
\end{aligned}
\end{equation}
where $\theta_t$ denotes the current policy parameters, $k$ indexes the $|y_t^{(j)}|$ tokens of rollout $j$, and $\hat{A}_t^{(j)}=(r_t^{(j)}-\mu(r_t))/(\sigma(r_t)+\epsilon_A)$ is the group-relative advantage. Here, $\mu(r_t)$ and $\sigma(r_t)$ are the group's reward mean and standard deviation, and $\epsilon_A$ is a small constant. The token-level importance ratio is $\rho_{j,k}=\policyt(y_{t,k}^{(j)}\mid x_t,\hctrl{g_t},y_{t,<k}^{(j)})\,/\,\pi_{\theta_{\mathrm{old}}}(\,\cdot\,)$ against $\pi_{\theta_{\mathrm{old}}}$, the policy that sampled the rollout. Finally, $\epsilon$ is the clip range, $\beta$ the KL coefficient, and $\mathbb{D}_{\mathrm{KL}}$ the KL divergence from the reference policy $\pi_{\mathrm{ref}}$.

\subsection{Harness Memory}
\label{sec:memory}

One step's score matrix shows the current pattern, but not whether a failure persists across steps or how earlier updates affected it. Harness memory $\mathcal{M}_t=\{m_\tau\}_{\tau\leq t}$ therefore keeps one record per task draw over the full run, organized as $m_t=\mtag{rollout history}\;\allowbreak\mtag{rubric history}\;\allowbreak\mtag{training dynamics}$. The \mtag{rollout history} field holds the task $x_t$ with its rollout group $Y_t$ and score matrix $S_t\in\{0,1\}^{K_t\times J}$, \mtag{rubric history} the harness $\harness{}_t$ with its proposed actions, and \mtag{training dynamics} the task's reward trajectory and update history. Because accumulated history soon exceeds the evolver's context, the evolver retrieves evidence $E_t=(E_t^{\mathrm{time}},E_t^{\mathrm{sem}})$ through two indexes keyed by the task and criterion text,
\begin{equation}
    E_t^{\mathrm{time}}=\{m_\tau\in\mathcal{M}_{t-1}: x_\tau=x_t\},
    \qquad
    E_t^{\mathrm{sem}}=\operatorname{top-}\!k\,\bigl\{(q',x'):\cos\bigl(\phi(q_{t,m}),\phi(q')\bigr)\geq\eta\bigr\} .
\end{equation}
The time index returns earlier task records, letting the attributor distinguish a persistent \clblind{} or \clsat{} failure from single-step noise and the proposer avoid repeating an already reversed update. The semantics index returns related cross-task criteria with positive and negative responses, supplying the $\gtag{reference}$ field when the task history has no such pair.

\subsection{Harness Evolver}
\label{sec:evolver}

The harness evolver $\mathcal{E}$ converts one step's evidence into the next harness $\harness{}_{t+1}$. Its three roles, the attributor, proposer, and JIT critic, separate attribution, proposal, and acceptance so no model call both proposes and accepts an update,
\begin{equation}
\begin{gathered}
    d_t = \role{attribute}\bigl(x_t,Y_t,S_t,\harness{}_t,E_t\bigr)\in\mathcal{V}^{K_t},\qquad
    a^{(r)} = \role{propose}\bigl(d_t,E_t,\harness{}_t,F^{(r-1)}\bigr)\in\mathcal{A},\\
    \bigl(\Delta^{(r)},F^{(r)}\bigr) = \role{critique}\bigl(\harness{}_t,\,a^{(r)}(\harness{}_t),\,\policyt,\,x_t,Y_t\bigr),
\end{gathered}
\end{equation}
where $a(\harness{})$ is the harness after action $a$ and $F^{(r)}$ records each update rejected at round $r$ with the measurement that rejected it, starting from $F^{(0)}=\emptyset$. The proposer and critic iterate until every candidate update is accepted or the round budget $B=3$ is exhausted. An unresolved interface keeps its current setting, and every label, action, and measurement is written to $m_t$.

\noindent\textbf{Harness attributor.} Scores establish whether a criterion separates the current group, but explaining its failure requires the task history in $E_t^{\mathrm{time}}$ and related criteria in $E_t^{\mathrm{sem}}$. Using both, the attributor asks three questions to assign each criterion a label in $\mathcal{V}=\{\clblind,\clsat,\clspur,\clagree\}$ and route each failure label to one interface. \textit{(i) Reach} asks whether the policy can produce the target property. A criterion unmet in the current group and earlier task records is labeled \clblind{} and routed to the rollout interface, whereas one unmet now but met earlier is treated as noise and labeled \clagree{}. \textit{(ii) Training stage} asks whether the policy has mastered the task. A criterion met by every rollout across recent steps is labeled \clsat{}. When such criteria lower the task's discriminative mass, the proportion of criteria separating the group, the task is diagnosed as \clsat{} and routed to the sampling interface. The attributor also flags a criterion met by every rollout through generic content, such as a disclaimer or polite closing, so the proposer can tighten it with a secondary rubric operation. \textit{(iii) Reward hacking} asks whether meeting a separating criterion reflects quality. The attributor compares rollouts that meet and fail the criterion with positive responses of related criteria retrieved by the semantics index. If only satisfying rollouts share superficial compliance, length gaming, partial coverage, or stance gaming through sycophancy or over-refusal, the criterion is labeled \clspur{} and routed to the rubric interface, and otherwise labeled \clagree{}. The attributor supports each label with quoted rollouts and raises its confidence as the label persists. A separate rubric-wording check identifies the exact phrase permitting a surface form or generic pass, which the attributor gives the proposer as the basis for a criterion rewrite.

\noindent\textbf{Harness proposer.} The label selects the interface, while the proposer selects its update. The action $a=(a_p,a_g,a_c)$ contains one candidate update or \texttt{noop} per interface, and the critic evaluates each update's typed operations together. Besides a label's primary update, the proposer may add secondary rubric operations on a \clsat{} or \clblind{} criterion to the rubric candidate or propose as a rollout update a \oguid{drop} of a field whose target has left \clblind{}. The sampling operation $\osamp{reweight}(i,\gamma_\ell)$ lowers task weight one level from $w^i_0=1$ through $\gamma_1\geq\gamma_2\geq\gamma_3$. The memory cold start (\Cref{sec:exp}) sets $\gamma_1$, $\gamma_2$, and $\gamma_3$ to task-tertile median discriminative masses. The guidance operations $\oguid{add}(f)$ and $\oguid{drop}(f)$ mount or remove a field $f\in\{\gtag{criterion},\gtag{reference},\gtag{critique}\}$ under a per-task cap that falls by one each training quarter. The rubric operations $\orubr{create}$, $\orubr{rewrite}$, $\orubr{delete}$, $\orubr{reweight}$, $\orubr{merge}$, and $\orubr{split}$ act on individual criteria~\citep{wu2026amaris}. Anchor criteria are seed criteria whose text the memory cold start fixes, so the proposer may only \orubr{reweight} them. Each rubric operation rescales positive weights to total $\Lambda$ while holding anchors at a fixed share to prevent influence drift.

\noindent\textbf{JIT critic.} Earlier evidence may support an update that is ineffective or excessive for the current policy, so the critic measures each primary candidate update just in time on that policy with parameters frozen before accepting it. Because a reward increase cannot validate a rubric update~\citep{chen2026decoevo}, a rubric update for a \clspur{} criterion must instead lower the recorded group's score. Holding the other two interfaces fixed, the critic regrades the recorded group for a rubric update, re-samples a group from the policy with parameters $\theta_t$ frozen under candidate guidance for a rollout update, and makes no policy query for a sampling update. With $\bar{r}(Y,c)=\frac{1}{J}\sum_{j=1}^{J}r(x_t,y^{(j)},c)$ denoting the group-average reward and $\bar{r}_t=\bar{r}(Y_t,c_t)$ its current value, the measurements are
\begin{equation}
    \Delta_c = \bar{r}_t - \bar{r}(Y_t,\widetilde{c}_{t+1}),\quad
    \Delta_g = \bar{r}(\widetilde{Y}_t,c_t) - \bar{r}_t,\quad
    \Delta_w = \bigl(w_t^{i}-\widetilde{w}_{t+1}^{i}\bigr)\,\bar{r}_t,
\end{equation}
where $\widetilde{Y}_t\sim\policyt(\cdot\mid x_t,\widetilde{g}_{t+1})$ is the re-sampled group, $\Delta_w$ is the task-weight reduction reallocated by sampling to other tasks, scaled by current group-average reward, and $u_t$ keeps the task's expected contribution to the objective unchanged. The critic accepts a candidate update when \textit{(i)} a sampling update for a \clsat{} task has $\Delta_w>0$, a rollout update for a \clblind{} criterion has $\Delta_g>0$, and a rubric update for a \clspur{} criterion has $\Delta_c>0$, \textit{(ii)} the $\Delta_g$ or $\Delta_c$ in \textit{(i)} lies within the corresponding acceptance band $[\varepsilon_{\mathrm{low}},\varepsilon_{\mathrm{high}}]$, a per-interface hyperparameter set by the memory cold start, \textit{(iii)} a rubric update for a \clspur{} criterion leaves every criterion whose text it creates or rewrites varying within the group, because a constant criterion contributes nothing to the policy update~\citep{guan2026evorubric,wang2026serpo}, and \textit{(iv)} a sampling update, which has no acceptance band, has a \clsat{} diagnosis in the task's previous time-index record. A joint rubric candidate containing secondary operations at a step with a \clspur{} label must pass conditions \textit{(ii)} and \textit{(iii)}. Otherwise, it is accepted with the step's primary update, and its $\Delta_c$ is recorded in $m_t$. The critic measures a \oguid{drop} as a rollout update on a group re-sampled without the field and accepts it when $\Delta_g\geq-\varepsilon^{g}_{\mathrm{low}}$. The accepted updates yield $\harness{}_{t+1}$, and rejected ones return with their measurements to the proposer for revision or replacement with \texttt{noop}.

\section{Experiments}
\label{sec:exp}

\begin{table}[t]
\centering
\footnotesize
\setlength{\tabcolsep}{3pt}
\renewcommand{\tabhead}[1]{{\scriptsize #1}}
\caption{%
Domain composition of \corpus{}. Each domain has $250$ held-out tasks. Token counts are means under the Qwen tokenizer. The Tasks (\%) column under Reference response reports the percentage of training tasks with a released reference response.}
\label{tab:corpus}
\begin{tabular*}{\textwidth}{@{\extracolsep{\fill}}lclcccccc@{}}
\tabtop
& & & \tabhead{Prompt} & \multicolumn{3}{c}{\tabhead{Seed rubric}} & \multicolumn{2}{c}{\tabhead{Reference response}} \\
\cmidrule(lr){4-4} \cmidrule(lr){5-7} \cmidrule(l){8-9}
\tabhead{Domain} & \tabhead{Train tasks} & \tabhead{OOD benchmark} & \tabhead{Tokens} & \tabhead{Criteria} & \tabhead{Tokens/crit.} & \tabhead{Negative (\%)} & \tabhead{Tokens} & \tabhead{Tasks (\%)} \\
\midrule
\tabsrc{Medical}{HealthBench, RaR, RubricHub} & 3{,}000 & \tabsub{MedQA}{250 items} & 121 & 12.4 & 43.4 & 21.0 & 322 & 67.7 \\
\tabsrc{Science}{ResearchQA, RaR, RubricHub} & 2{,}000 & \tabsub{GPQA-Diamond}{198 items} & 50 & 12.3 & 31.8 & 2.0 & 362 & 31.1 \\
\tabsrc{Writing}{LongWriter-Zero} & 1{,}200 & \tabsub{WritingBench}{250 items} & 161 & 10.5 & 21.6 & 28.1 & -- & 0.0 \\
\tabsrc{Role-play}{RubricHub (chat)} & 1{,}200 & \tabsub{RoleBench}{250 items} & 85 & 26.5 & 31.5 & 0.0 & 1{,}205 & 24.2 \\
\tabsrc{Instruction following}{RubricHub (IF)} & 1{,}200 & \tabsub{IFBench}{250 items} & 211 & 3.6 & 52.0 & 0.0 & 528 & 5.6 \\
\midrule
All & 8{,}600 & & 118 & 12.9 & 35.2 & 10.7 & 420 & 35.0 \\
\tabbottom
\end{tabular*}
\end{table}

\noindent\textbf{Dataset.} To evaluate \ourmethod{}'s effectiveness and efficiency, we propose \corpus{}, curated from five public rubric-annotated sources across five open-ended domains (\Cref{tab:corpus}). Medical tasks come from HealthBench~\citep{arora2025healthbench} and the medical slices of RaR~\citep{gunjal2025rar} and RubricHub~\citep{li2026rubrichub}, science tasks from ResearchQA~\citep{yifei2025researchqa} and the science slices of RaR and RubricHub, writing tasks from LongWriter-Zero~\citep{wu2025longwriterzero}, and role-play and instruction-following tasks from the chat and instruction slices of RubricHub. For each source, we drop ineligible tasks and overlong prompts before drawing a fixed quota. This curation yields $8{,}600$ training tasks with seed rubrics of weighted criteria, $250$ held-out tasks per domain disjoint from training, and one OOD benchmark per domain. The training tasks contain $110{,}606$ criteria, averaging $12.9$ criteria per task with a median of $10$. Prompts average $118$ Qwen tokens with a median of $54$, while criteria average $35.2$ Qwen tokens. Negative-weight criteria constitute $10.7\%$ of all criteria and appear in $48.6\%$ of tasks. Released reference responses accompany $3{,}008$ training tasks ($35.0\%$). These responses comprise physician-written ideal completions for HealthBench, reference answers for RaR, and the highest-scoring of six samples for RubricHub. They average $420$ tokens, with domain means from $322$ in medical to $1{,}205$ in role-play. ResearchQA and LongWriter-Zero provide none. Writing prompts specify a target length range. We use the released references only to describe the corpus, never in training or the reward.

\noindent\textbf{Evaluation.} The frozen training judge scores in-domain held-out tasks under their seed rubrics, unchanged by harness updates. The official OOD protocols use multiple-choice accuracy for MedQA~\citep{jin2020medqa} and GPQA-Diamond~\citep{rein2023gpqa_fix}, ROUGE-L for RoleBench~\citep{wang-etal-2024-rolellm}, programmatic constraint checks for IFBench~\citep{pyatkin2025ifbench}, and WritingBench's own rubric~\citep{wu2025writingbench}. Scores are comparable only within a benchmark. Because only WritingBench uses the training judge, the other four benchmarks test whether gains persist beyond the harness-updated reward. We decode greedily without guidance and evaluate OOD every $25$ steps and in domain every~$100$. Every results table reports the seed mean at the best checkpoint of the seed-mean curve for each method and evaluation set. Each Avg.\ column is the unweighted mean of the five domain columns.

\begin{table}[t]
\centering
\footnotesize
\setlength{\tabcolsep}{1pt}
\caption{%
Quantitative results of \ourmethod{} on the five-domain \corpus{} with \mbox{Qwen3.5-4B} and Qwen3.5-9B.
\textbf{Bold}, \underline{underline}, and {\color{gaingreen}green} mark the best result, the second-best result, and \ourmethod{}'s margin over the best baseline, respectively.}
\label{tab:main}
\begin{tabularx}{\textwidth}{@{}l*{6}{Y}*{6}{Y}@{}}
\tabtop
& \multicolumn{6}{c}{\tabhead{In-domain held-out}} & \multicolumn{6}{c}{\tabhead{Out of distribution}} \\
\cmidrule(lr){2-7} \cmidrule(l){8-13}
\tabhead{Method} & \tabhead{Med.} & \tabhead{Sci.} & \tabhead{Writ.} & \tabhead{Role} & \tabhead{IF} & \tabhead{Avg.} & \tabhead{Med.} & \tabhead{Sci.} & \tabhead{Writ.} & \tabhead{Role} & \tabhead{IF} & \tabhead{Avg.} \\
\midrule
\multicolumn{13}{l}{\tabgroup{Qwen3.5-4B}} \\
GRPO & 52.4 & 57.2 & 60.7 & 50.8 & 68.1 & 57.8 & 40.5 & 29.3 & 66.5 & 50.7 & 32.7 & 43.9 \\
\addlinespace[1.6pt]
RuscaRL & \snd{56.3} & 61.4 & 63.1 & 53.6 & 69.5 & 60.8 & 41.3 & 29.8 & 67.1 & 51.1 & 32.2 & 44.3 \\
SkillRL & 55.6 & 60.9 & 62.4 & 53.1 & \snd{69.8} & 60.4 & 41.0 & 30.1 & 67.6 & 51.4 & 32.9 & 44.6 \\
\addlinespace[1.6pt]
EvoRubric & 55.0 & 60.5 & 64.8 & 54.2 & 69.1 & 60.7 & 43.4 & 31.2 & 69.1 & 53.1 & \snd{34.3} & 46.2 \\
AMARIS & 54.7 & 60.2 & 65.3 & \snd{55.2} & 69.2 & 60.9 & 43.9 & 31.6 & 69.8 & 53.5 & 33.9 & 46.5 \\
\addlinespace[1.6pt]
AMARIS++ & 56.1 & \snd{61.7} & \snd{65.5} & 55.0 & 69.4 & \snd{61.5} & \snd{44.1} & \snd{32.3} & \snd{70.6} & \snd{54.0} & 34.1 & \snd{47.0} \\
\addlinespace[1.6pt]
\rowcolor{oursbg} \ourmethod{} & \gain{59.3}{3.0} & \gain{64.6}{2.9} & \gain{69.2}{3.7} & \gain{58.9}{3.7} & \gain{70.7}{0.9} & \gain{64.5}{3.0} & \gain{46.5}{2.4} & \gain{35.4}{3.1} & \gain{73.0}{2.4} & \gain{56.6}{2.6} & \gain{37.5}{3.2} & \gain{49.8}{2.8} \\
\midrule
\multicolumn{13}{l}{\tabgroup{Qwen3.5-9B}} \\
GRPO & 58.9 & 63.4 & 65.8 & 56.2 & 73.5 & 63.6 & 52.6 & 35.5 & 71.8 & 58.4 & 38.9 & 51.4 \\
\addlinespace[1.6pt]
RuscaRL & \snd{61.9} & 66.8 & 67.4 & 58.4 & 74.6 & 65.8 & 53.3 & 35.9 & 72.1 & 58.7 & 38.6 & 51.7 \\
SkillRL & 61.2 & 66.1 & 67.0 & 58.0 & \snd{74.9} & 65.4 & 53.0 & 36.2 & 72.5 & 58.9 & 39.2 & 52.0 \\
\addlinespace[1.6pt]
EvoRubric & 60.8 & 65.9 & 69.3 & 59.4 & 74.2 & 65.9 & 55.1 & 37.4 & 74.0 & 60.3 & \snd{40.6} & 53.5 \\
AMARIS & 60.5 & 65.6 & 69.8 & \snd{59.9} & 74.3 & 66.0 & 55.6 & 37.8 & 74.6 & 60.8 & 40.2 & 53.8 \\
\addlinespace[1.6pt]
AMARIS++ & 61.7 & \snd{67.0} & \snd{70.4} & 59.6 & 74.4 & \snd{66.6} & \snd{56.0} & \snd{38.4} & \snd{75.2} & \snd{61.2} & 40.1 & \snd{54.2} \\
\addlinespace[1.6pt]
\rowcolor{oursbg} \ourmethod{} & \gain{64.6}{2.7} & \gain{69.9}{2.9} & \gain{73.8}{3.4} & \gain{63.5}{3.6} & \gain{75.9}{1.0} & \gain{69.5}{2.9} & \gain{58.9}{2.9} & \gain{41.6}{3.2} & \gain{78.3}{3.1} & \gain{63.9}{2.7} & \gain{43.6}{3.0} & \gain{57.3}{3.1} \\
\tabbottom
\end{tabularx}
\vspace{-3mm}
\end{table}

\noindent\textbf{Baselines.} We compare static-rubric GRPO~\citep{shao2024deepseekmath_cased}, guidance-side RuscaRL~\citep{zhou2025ruscarl} and SkillRL~\citep{xia2026skillrl_fix} for scaffolded rollouts over a frozen rubric, and rubric-side EvoRubric~\citep{guan2026evorubric} and AMARIS~\citep{wu2026amaris} for rubric rewriting without guidance. AMARIS++, the baseline closest to \ourmethod{}, extends AMARIS to all three interfaces but lacks cross-task memory and a JIT critic. All arms share the policy, judge, seed rubrics, decoding, and $500$-step budget and differ only in whether and how they adapt the harness interfaces.

\noindent\textbf{Training configuration.} We train Qwen3.5-4B and Qwen3.5-9B~\citep{qwen35_2026} with GRPO for $500$ steps, drawing $512$ tasks per step and sampling $J=8$ rollouts per task at temperature $1.0$ and top-$p$ $0.9$. We use AdamW at learning rate $1\times10^{-6}$, and the objective uses KL coefficient $\beta=0.001$, clip range $\epsilon=0.2$, and advantage constant $\epsilon_A=10^{-6}$. Every arm trains with three seeds. Before training, one memory cold start runs the Qwen3.5-4B seed policy over the entire training set pooled across the five domains, and the Qwen3.5-9B runs reuse its values. Setting $\Lambda=100$ makes $\Lambda\bar{r}$ the group's mean score in reward points, the unit of every $\Delta$ and acceptance band. The cold start sets the sampling levels $\gamma_1=0.61$, $\gamma_2=0.38$, and $\gamma_3=0.17$, and the rollout and rubric acceptance bands on group-average reward changes to $[10.2,28.0]$ and $[3.1,28.0]$ points. The rollout and rubric lower bounds are the $95$th percentiles of null variation across independent groups of one task and repeated gradings of one group, respectively. Both upper bounds are the median within-group reward range.

\noindent\textbf{Implementation.} Frozen \mbox{gemma-4-31B-it}~\citep{gemmateam2026gemma4} serves as the judge at temperature $0.6$ and as the backbone of all three evolver roles. ChromaDB stores the time index, keyed by task, and the semantics index, which embeds criterion text with bge-base-en-v1.5 as $\phi$ and returns the top-$k=4$ criteria whose cosine similarity is at least the floor $\eta=0.82$. The critic rejects a candidate update when its call returns no measurement. Costs are in cluster hours, the wall-clock hours of one $16$-node cluster with eight NVIDIA H100 80GB GPUs per node.

\noindent\textbf{Oracle judge.} For analysis only, Claude Sonnet 5 is the oracle judge. It is stronger than the training judge, reads no rubric, and compares response pairs under one quality guideline per domain, judging clinical correctness and safety in medicine, scientific correctness and reasoning in science, fit to purpose and genre in writing, persona consistency in role-play, and constraint satisfaction in instruction following. For each group of $J=8$ rollouts, the oracle scores all $28$ pairs in both orders and averages each pair's scores (win $1$, tie $0.5$). Rollout win counts define the within-group oracle ranking, with equal counts tied, against which we compare each pair's reward order. A tied-reward pair is \clblind{} when the shared reward is below $50$ points and \clsat{} otherwise, and a reward-separated pair is \clagree{} when its reward order matches the oracle order and \clspur{} otherwise. Oracle quality is $100$ times a response's normalized win rate against eight fixed \mbox{gemma-4-31B-it} comparison responses per task. On $240$ sampled groups, two annotators agree at Kendall's $\tau=0.84$, and the oracle ranking agrees with each annotator at $0.80$ and $0.79$ and with their consensus at $0.82$. No harness role sees oracle output.

\begin{table}[t]
\centering
\footnotesize
\setlength{\tabcolsep}{0pt}
\caption{%
Ablations of \ourmethod{} on its components and control interfaces at Qwen3.5-4B.
\textbf{Bold} and \underline{underline} mark the best and second-best results, respectively.}
\label{tab:ablation}
\begin{tabularx}{\textwidth}{@{}l*{6}{Y}*{6}{Y}@{}}
\tabtop
& \multicolumn{6}{c}{\tabhead{In-domain held-out}} & \multicolumn{6}{c}{\tabhead{Out of distribution}} \\
\cmidrule(lr){2-7} \cmidrule(l){8-13}
\tabhead{Method} & \tabhead{Med.} & \tabhead{Sci.} & \tabhead{Writ.} & \tabhead{Role} & \tabhead{IF} & \tabhead{Avg.} & \tabhead{Med.} & \tabhead{Sci.} & \tabhead{Writ.} & \tabhead{Role} & \tabhead{IF} & \tabhead{Avg.} \\
\midrule
\rowcolor{oursbg} \ourmethod{} & \best{59.3} & \best{64.6} & \best{69.2} & \best{58.9} & \best{70.7} & \best{64.5} & \best{46.5} & \best{35.4} & \best{73.0} & \best{56.6} & \best{37.5} & \best{49.8} \\
\addlinespace[2pt]
\multicolumn{13}{l}{\tabgroup{Component ablation}} \\
\quad w/o critic & 54.9 & 60.0 & 62.1 & 52.0 & 68.3 & 59.5 & 42.9 & 31.6 & 68.5 & 52.7 & 33.8 & 45.9 \\
\quad w/o critic \& memory & 53.6 & 58.6 & 60.8 & 50.7 & 67.0 & 58.1 & 41.3 & 29.2 & 66.9 & 51.0 & 32.2 & 44.1 \\
\addlinespace[2pt]
\multicolumn{13}{l}{\tabgroup{Interface ablation}} \\
\quad \ourmethod{}$_g$ & 56.1 & 61.1 & 63.2 & 53.1 & 69.5 & 60.6 & 43.4 & 31.7 & 69.0 & 53.0 & 34.3 & 46.3 \\
\quad \ourmethod{}$_c$ & 56.7 & 61.7 & 63.8 & 53.7 & 70.1 & 61.2 & 44.3 & 32.6 & 69.9 & 54.0 & 35.2 & 47.2 \\
\addlinespace[2pt]
\quad $p_t\leftarrow$ SEC & \snd{58.6} & \snd{63.9} & \snd{68.5} & \snd{58.1} & \snd{70.3} & \snd{63.9} & \snd{45.9} & \snd{34.8} & \snd{72.3} & \snd{56.0} & \snd{36.9} & \snd{49.2} \\
\quad $g_t\leftarrow$ RuscaRL & 58.4 & 63.7 & 67.9 & 57.6 & 70.2 & 63.6 & 44.9 & 33.6 & 71.2 & 54.9 & 36.1 & 48.1 \\
\quad $c_t\leftarrow$ EvoRubric & 58.0 & 63.2 & 68.0 & 57.5 & 69.9 & 63.3 & 45.1 & 33.9 & 71.6 & 55.2 & 36.3 & 48.4 \\
\tabbottom
\end{tabularx}
\vspace{-3mm}
\end{table}

\subsection{Main Results}
\label{sec:main}

\textbf{\ourmethod{} leads every benchmark at both policy scales.} At Qwen3.5-4B, \ourmethod{} reaches $64.5$ in domain and $49.8$ OOD (\Cref{tab:main}), $+6.7$ and $+5.9$\,pp above GRPO and $+3.0$ and $+2.8$\,pp above AMARIS++, the strongest baseline. At Qwen3.5-9B, it reaches $69.5$ and $57.3$, $+2.9$ and $+3.1$\,pp above AMARIS++, which it exceeds on every seed at both scales. Paired bootstrap $95\%$ intervals for the AMARIS++ margins ($10{,}000$ item-and-seed resamples) are $[2.4,3.6]$ in domain and $[2.0,3.5]$ OOD at Qwen3.5-4B, and $[2.2,3.6]$ in domain and $[2.3,3.9]$ OOD at Qwen3.5-9B. Paired permutation tests give $p<0.001$ against AMARIS++ and GRPO. Seed standard deviations are below one point for all arms and smallest for \ourmethod{} ($0.3$ and $0.4$ at Qwen3.5-4B, $0.2$ and $0.3$ at Qwen3.5-9B). We attribute this stability to the critic rejecting rollout and rubric updates whose measured effects leave the acceptance bands.

\textbf{\ourmethod{}'s in-domain margin over the best baseline is widest on writing and role-play and narrowest on instruction following.} At Qwen3.5-4B, it reaches $+3.7$\,pp on writing and role-play but falls to $+0.9$\,pp on instruction following. We attribute the wide margin to many defensible responses saturating a frozen rubric and leaving further gains to rubric updates measured on the recorded group, and the narrow one to near-verifiable criteria on which GRPO already reaches $68.1$.

\textbf{Guidance-side and rubric-side adaptation differ in OOD transfer.} At Qwen3.5-4B, guidance-side RuscaRL and SkillRL gain $3.0$ and $2.6$\,pp over GRPO in domain ($60.8$ and $60.4$ against $57.8$) but barely transfer OOD ($44.3$ and $44.6$ against $43.9$). Rubric-side EvoRubric and AMARIS match this in-domain gain ($60.7$ and $60.9$) and transfer further OOD ($46.2$ and $46.5$). We attribute limited guidance-side OOD transfer to scaffolds fitting the policy to the frozen training rubric, and stronger rubric-side transfer to a rewritten criterion changing what the reward measures. Three-interface AMARIS++ is the strongest baseline at both scales. Harness memory and the critic lift \ourmethod{} from $44.1$ to $49.8$ OOD (\Cref{tab:ablation}).

\subsection{In-Depth Analysis}
\label{sec:signal}

\begin{figure}[t]
\centering
\includegraphics[width=\textwidth]{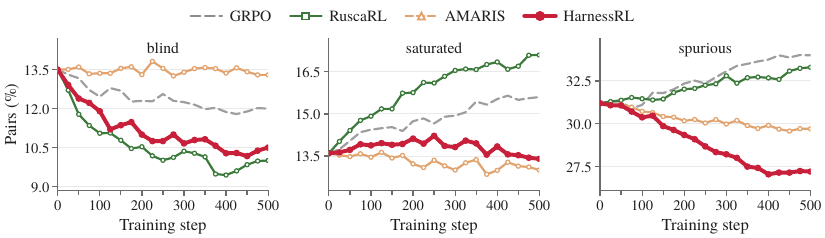}
\caption{%
Proportion of pairs by failure class at Qwen3.5-4B during training.}
\label{fig:dynamics}
\vspace{-3mm}
\end{figure}

\noindent\textbf{RQ1: What does each interface and component contribute?}\ \Cref{tab:ablation} reports best-checkpoint seed means in domain and OOD at Qwen3.5-4B. The component ablation removes the critic (\emph{w/o critic}) and then memory retrieval (\emph{w/o critic \& memory}). Interface ablations include single-interface harnesses adapting only the rollout or rubric interface (\ourmethod{}$_g$, \ourmethod{}$_c$) and switching arms using a published rule for one interface. Each switching arm retains harness memory and the harness evolver for two interfaces, while the third uses SEC's difficulty bandit ($p_t\leftarrow$ SEC~\citep{chen2025sec}), RuscaRL's scheduled checklist scaffold ($g_t\leftarrow$ RuscaRL~\citep{zhou2025ruscarl}), or EvoRubric's criterion pruning and regeneration ($c_t\leftarrow$ EvoRubric~\citep{guan2026evorubric}), with no attributor, proposer, or critic.

\textbf{Single-interface harnesses trail \ourmethod{}, whose gain over GRPO comes mostly from the critic with memory.} \ourmethod{}$_g$ and \ourmethod{}$_c$ reach $60.6$ and $61.2$ in domain and $46.3$ and $47.2$ OOD, against $64.5$ and $49.8$ for \ourmethod{}. Removing the critic leaves $59.5$ in domain and $45.9$ OOD, below AMARIS++ ($47.0$). Removing memory too leaves $58.1$ in domain and $44.1$ OOD, within $0.2$\,pp of GRPO. We attribute this to memory reducing the reversal rate (RQ2) and the critic measuring effects before acceptance.

\textbf{A published rule on one interface keeps most of the gain over GRPO but never all of it.} The sampling, rubric, and rollout switching arms reach $63.9$, $63.3$, and $63.6$ in domain and $49.2$, $48.4$, and $48.1$ OOD, above both single-interface harnesses and AMARIS++ ($47.0$ OOD, $61.5$ in domain). Relative to \ourmethod{}, the rollout switching arm's OOD shortfall is largest at $1.7$\,pp and is $0.9$\,pp in domain, while the rubric and sampling switching arms trail by $1.4$ and $0.6$\,pp OOD. We attribute the sampling shortfall to reallocation leaving every rollout's reward unchanged, the rubric shortfall to rubric updates the critic never measures, and the rollout shortfall to scheduled scaffolds teaching the policy the training rubric, as RuscaRL's in-domain gain barely transfers OOD in \Cref{tab:main}.

\begin{figure}[t]
\centering
\includegraphics[width=\textwidth]{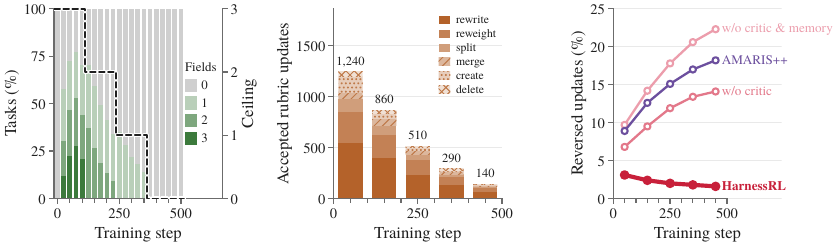}
\caption{%
Training dynamics of \ourmethod{}: proportion of tasks by guidance fields, accepted rubric updates per 100 steps, and reversed rubric updates, the last also for its ablations and AMARIS++.}
\label{fig:harness}
\vspace{-3mm}
\end{figure}

\noindent\textbf{RQ2: Why must the three interfaces adapt jointly, and what stabilizes the accepted updates?}\ At Qwen3.5-4B, we compare the arms at step $500$. \Cref{fig:dynamics} plots \clblind{}, \clsat{}, and \clspur{} proportions for GRPO, RuscaRL, AMARIS, and \ourmethod{}, and we report the \clagree{} proportion as reward-oracle agreement. Two arms used only here apply \emph{w/o critic \& memory} to \ourmethod{}$_g$ and \ourmethod{}$_c$. Spearman correlations relate the step-$500$ proportions, negated except for \clagree{}, to best-checkpoint scores across twelve arms, those in \Cref{tab:main} and \Cref{tab:ablation} except SkillRL and the switching arms, plus these two. From memory records, we report the mounted guidance fields per task, the proportion of tasks at each sampling level, the donors for \gtag{reference} fields mounted by accepted rollout updates, and the critic's acceptance rate with and without retrieved evidence above the floor $\eta$. We report the reversal rate in \Cref{fig:harness}, defined as the proportion of accepted rubric updates in each $100$-step window that undo an earlier same-criterion update within the task's previous two draws. Three annotators, blind to step and rationale, judged $200$ attributor labels ($50$ per class) and $200$ accepted rubric updates sampled uniformly from \ourmethod{}, with mean pairwise Cohen's $\kappa$ of $0.71$ and $0.68$. We aggregate judgments by majority vote.

\textbf{Guidance-side and rubric-side arms trade one failure class for another, whereas three-interface arms with memory lower all three, \ourmethod{} the most.} GRPO ends at $12.0$\slash$15.6$\slash$34.0$ for \clblind{}, \clsat{}, and \clspur{}. RuscaRL lowers \clblind{} to $10.0$ but raises \clsat{} to $17.1$, while \ourmethod{}$_g$ raises \clsat{} to $16.1$, or $16.5$ without critic and memory. AMARIS lowers \clsat{} and \clspur{} to $13.0$ and $29.7$ but raises \clblind{} to $13.3$, while \ourmethod{}$_c$ raises \clblind{} to $13.2$, or $13.7$ without critic and memory. \ourmethod{} without the critic reaches $11.1$\slash$15.2$\slash$31.6$, and removing memory too raises \clsat{} to $16.1$. AMARIS++, whose memory is a per-task revision log, reaches $11.4$\slash$14.6$\slash$30.8$, and \ourmethod{} reaches $10.5$\slash$13.4$\slash$27.2$, with reward-oracle agreement at $48.9$ against GRPO's $38.4$. We attribute the \clsat{} rise to guidance saturating a fixed rubric~\citep{yu2026dapo}, and the \clblind{} rise to rubric tightening without guidance placing criteria beyond the policy's reach~\citep{sundaram2026learnability}.

\textbf{The \clagree{}, \clspur{}, and \clsat{} proportions rank the arms, but the \clblind{} proportion does not.} In domain/OOD, correlations are $0.91/0.92$ for \clagree{}, $0.78/0.82$ for \clspur{}, $0.59/0.74$ for \clsat{}, and $0.15/{-}0.01$ for \clblind{}. We attribute the weak \clblind{} correlation to this trade, as rubric-side arms have the highest \clblind{} proportions yet score higher OOD than GRPO and the guidance-side baselines (\Cref{tab:main}).

\textbf{Guidance is withdrawn ahead of schedule, rubric updates concentrate early, and sampling down-weights half the tasks.} Mounted guidance fields average $1.6$ per task at their step-$75$ peak, whereas without the critic the count stays near the cap until each reduction. By step $350$, before the last reduction at step $375$, $86\%$ of tasks carry no guidance fields. Accepted rubric updates fall from $1{,}240$ in the first window to $140$ in the last, with \orubr{create} concentrated early. By step $500$, $52\%$ of tasks are below full weight, $8\%$ at the lowest level. We attribute the early withdrawal to the critic accepting a \oguid{drop} once its target leaves \clblind{}, and the decline of \orubr{create} to the time index showing the proposer existing criteria to revise.

\textbf{The critic with memory keeps reversed rubric updates rare.} The reversal rate ends at $1.6\%$ for \ourmethod{} and $18.2\%$ for AMARIS++, compared with $22.3\%$ without critic and memory. Without the critic, the reversal rate stays below that level but rises from $6.8\%$ in the first window to $14.1\%$ in the last. We attribute this rise to the missing critic, because memory shows the proposer earlier reversals but only the critic rejects a repeated update. Harness memory also supplies cross-task donors for $58\%$ of the \gtag{reference} fields mounted by accepted rollout updates. The critic accepts $71\%$ of candidate updates grounded in retrieved evidence, compared with $44\%$ when retrieval returns nothing above the floor.

\textbf{Annotators confirm attributor labels and accepted rubric updates.} Majority vote upheld the \clblind{}, \clsat{}, \clspur{}, and \clagree{} labels in $97.0\%$, $98.0\%$, $86.0\%$, and $90.0\%$ of cases, respectively, and judged $82.5\%$ of accepted rubric updates to address the diagnosed failure.

\begin{figure}[t]
\centering
\includegraphics[width=0.9\textwidth]{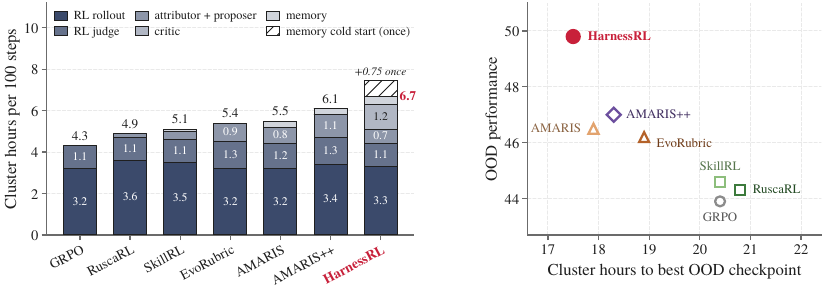}
\caption{%
Training efficiency of \ourmethod{} measured by cluster hours per 100 steps, decomposed into RL loop and harness evolver overhead (left), and cluster hours to the best OOD checkpoint (right).}
\label{fig:efficiency}
\vspace{-5mm}
\end{figure}

\noindent\textbf{RQ3: Does the harness recover its training overhead?}\ For every arm of \Cref{tab:main}, we measure Qwen3.5-4B training cost in cluster hours per 100 steps and identify its best OOD step at both scales. We decompose cost into RL rollout, RL judge, attributor and proposer, critic, and memory, and separately report the one-off memory cold start of $0.75$ cluster hours (\Cref{fig:efficiency}, left). We pair each arm's best OOD score with the cluster hours required to reach it, including the memory cold start (\Cref{fig:efficiency}, right). OOD scores peak before step $500$ for every arm, whereas in-domain scores rise through step $500$ for GRPO and every rubric-evolving method. The tables therefore select the best checkpoint for each evaluation set. To account for selection on the evaluation sets, we also report each arm's step-$500$ checkpoint. For matched compute, we report each arm's best OOD score among checkpoints evaluated every $25$ steps within the cluster hours, including the memory cold start, that \ourmethod{} needs to reach its best OOD checkpoint.

\textbf{At $1.6\times$ GRPO's per-step cost, \ourmethod{} reaches its best OOD checkpoint in fewer cluster hours than any baseline.} GRPO costs $4.3$ cluster hours per 100 steps versus $6.7$ for \ourmethod{}. The $2.4$ difference comprises $1.2$ for the critic, $0.7$ for attributor and proposer calls, $0.4$ for memory, and $0.1$ for longer guided rollouts ($3.3$ versus $3.2$). \ourmethod{} peaks at step $250$ after $17.5$ cluster hours. AMARIS, next in cluster hours, peaks at step $325$ after $17.9$, AMARIS++ at step $300$ after $18.3$, and GRPO at step $475$ after $20.4$ (\Cref{fig:efficiency}, right). EvoRubric, SkillRL, and RuscaRL peak at steps $350$, $400$, and $425$, respectively. At Qwen3.5-9B, the peak-step order is unchanged, with \ourmethod{} at step $250$, AMARIS++ at $325$, and GRPO at $475$. We attribute the earlier peak to the harness correcting the reward signal as it degrades and to the critic. GRPO makes no such correction, and the rubric-evolving baselines lack the critic. By mid-training, the critic has accepted the withdrawal of most guidance, and accepted rubric updates have declined from $1{,}240$ to $510$ per 100 steps.

\textbf{\ourmethod{} remains strongest without best-checkpoint selection and leads AMARIS++ under matched compute.} At step $500$, in-domain/OOD scores are GRPO $57.8/43.8$, RuscaRL $60.5/44.2$, SkillRL $60.1/44.5$, EvoRubric $60.7/45.7$, AMARIS $60.9/46.0$, AMARIS++ $61.5/46.4$, and \ourmethod{} $64.5/49.4$. Without best-checkpoint selection, \ourmethod{} exceeds AMARIS++ by $+3.0$\,pp on both evaluation sets. Within $17.5$ cluster hours, GRPO reaches $43.8$ OOD at step $400$, AMARIS++ reaches $46.9$ at step $225$, and \ourmethod{} reaches $49.8$ at step $250$. The per-step overhead therefore does not reduce the OOD lead over AMARIS++, which is $+2.9$\,pp under matched compute versus $+2.8$\,pp at best checkpoints.

\begin{figure}[t]
\centering
\includegraphics[width=\textwidth]{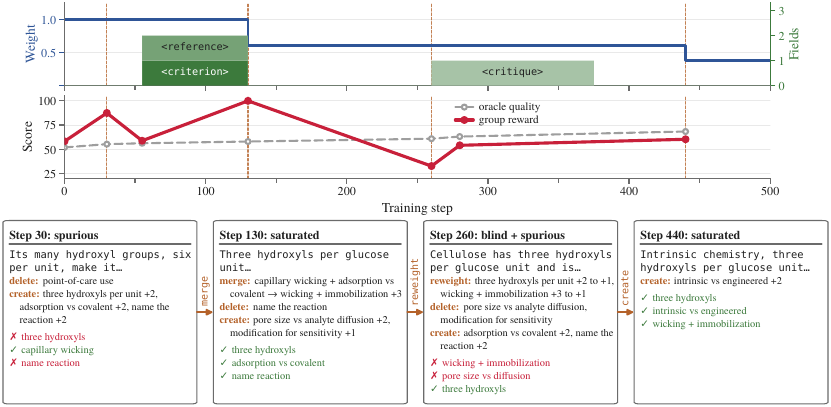}
\caption{%
Case analysis of a randomly sampled training task \emph{``Which properties of cellulose filter paper suit it to analytical and clinical use?''}: task sampling and mounted guidance (top), rollout groups scored with the dynamically evolving rubric and oracle quality (middle), and steps at which the harness acted, with labels, rollout excerpts, rubric operations, and criteria met or missed (bottom). Dashed lines mark them.}
\label{fig:case}
\vspace{-5mm}
\end{figure}

\noindent\textbf{RQ4: How does the harness act on one task over training?}\ \Cref{fig:case} traces a randomly sampled training task on cellulose filter paper, recording group reward under the current rubric and oracle quality whenever the harness acts. Because its seed criteria are not anchor criteria, the proposer may apply \orubr{delete} or \orubr{merge} to them.

\textbf{Each accepted primary update addresses the failure labeled at its step, and group reward diverges from oracle quality.} At step $30$, a rubric update for a \clspur{} label replaces a seed criterion with hydroxyl-count and coupling-chemistry criteria. At $55$, no rollout names the coupling reaction, so a rollout update for the resulting \clblind{} label mounts \gtag{criterion} and \gtag{reference} fields. A rollout update drops them at $130$ once their target criterion is met. At $130$, a sampling update for a \clsat{} label lowers the task weight to $\gamma_1$, while secondary rubric operations merge the wicking and adsorption criteria and delete the reaction-name criterion. At $260$, a rollout update for a \clblind{} label mounts a \gtag{critique} field, while a rubric update for a \clspur{} label deletes the sensitivity criterion and restores both coupling-chemistry criteria. At $280$, a rubric update for a \clspur{} label on the reaction-name criterion lowers both coupling-chemistry criteria to $+1$ and creates a filter-grade criterion at $+2$. At $440$, the attributor labels the hydroxyl-count criterion \clsat{} and all others \clagree{}. A sampling update lowers the task weight to $\gamma_2$, and a secondary rubric operation creates an intrinsic-versus-engineered criterion. From $130$ to $260$, group reward falls from $100.0$ to $32.8$ while oracle quality rises from $58.1$ to $61.0$. We attribute the fall to the changed rubric. Because the rubric and policy both affect group reward across steps, the critic compares the current and candidate rubrics on one recorded group at fixed $\theta$.

\section{Related work}
\label{sec:related}

\noindent\textbf{Evolving rubric-based RL.} Rubric-based RL rewards open-ended generation with natural-language criteria~\citep{gunjal2025rar,huang2025rubicon,viswanathan2025rlcf,liu2025openrubrics,li2026rubrichub,chen2026rubricsurvey}, and reward hacking against fixed rubrics~\citep{mahmoud2026rubrichacking,wang2026rubrichacking} motivates criteria and task distributions that evolve with the policy.
One line rewrites the rubric from the policy's responses~\citep{guan2026evorubric,wu2026amaris,wu2026dynamicrubric,wang2026serpo,chen2026decoevo}. EvoRubric~\citep{guan2026evorubric} prunes and regenerates criteria for open-ended generation, and AMARIS~\citep{wu2026amaris} revises the rubric through a memory-augmented loop.
Another line co-evolves a rubric generator with the judge or the policy, adapts the rubric to each comparison, or orders rubrics as a curriculum~\citep{xu2026rubricarm,li2026evolm,jia2026openrs,krishnan2026rcrl}. Rubric-ARM~\citep{xu2026rubricarm} alternates RL between a rubric generator and the judge that scores under its rubrics, and OpenRS~\citep{jia2026openrs} builds a rubric for each response pair.
A third line adapts the task distribution using priority, difficulty, or expected learning gain~\citep{schaul2016prioritized,gao2025pcl,sundaram2026learnability,xu2026pac,zeng2026cures}, and SEC~\citep{chen2025sec} evolves the curriculum with the policy.
\ourmethod{} instead adapts task sampling, rollout guidance, and rubric criteria in one training-time harness. It routes each attributed failure class to the interface whose update addresses it.

\noindent\textbf{Scaffolded and privileged rollouts.} Scaffolded rollouts change the composition of the sampled group so that it contains rewarded responses~\citep{yu2026dapo,cui2025entropy,zhou2025ruscarl,qu2026pope}, whereas rubric evolution changes the criteria that define the reward.
One line shapes the rollout group without changing the input~\citep{yu2026dapo,cui2025entropy}. DAPO~\citep{yu2026dapo} resamples groups whose rewards carry no advantage, and entropy control~\citep{cui2025entropy} keeps policy entropy from collapsing.
Another line places privileged context in the policy's input~\citep{zhou2025ruscarl,xia2026skillrl_fix,huang2026ocgrpo,li2025guide,hou2026riserl,qu2026pope,yu2026knowrl}. RuscaRL~\citep{zhou2025ruscarl} scaffolds exploration with the rubric and fades the scaffold on a schedule, and SkillRL~\citep{xia2026skillrl_fix} augments rollouts with recursively accumulated skills.
More recently, self-distillation ties guided exploration to the unguided policy~\citep{zhao2026opsd,dilkes2026ice}. OPSD~\citep{zhao2026opsd} applies on-policy self-distillation to guided reasoning, and ICE~\citep{dilkes2026ice} distills instruction-conditioned exploration into the unconditioned policy.
\ourmethod{} mounts a guidance field only after the critic measures a reward increase on a re-sampled group. It withdraws the field earlier than its own schedule requires, once the target criterion leaves \clblind{} and reward stays within tolerance without it.

\noindent\textbf{Training-time harness.} Agent harnesses usually denote the inference-time infrastructure that connects a model to its environment, such as interfaces, scaffolds, and memory~\citep{wang-etal-2025-r3mem,wang2026latentharness,gong2026grove}, and harness research~\citep{zhou2026externalization_fix,liu2025advances} treats them as one configuration to design and learn. Beyond inference, recent work brings the harness into training.
One line designs inference-time harnesses for frozen models~\citep{yang2024sweagent,suzgun2024meta,wang2025oscar}. SWE-agent~\citep{yang2024sweagent} gives the model an agent-computer interface for software engineering, and Meta-Prompting~\citep{suzgun2024meta} scaffolds the model with task-agnostic prompts.
Another line evolves harnesses from agent trajectories~\citep{ning2026evoharness,ning2026harnessbank,shi2026evolving,wang2026memcodex}. EvoHarness-RL~\citep{ning2026evoharness} learns a self-evolving runtime harness for long-horizon agents, and HarnessBank~\citep{ning2026harnessbank} searches a gene bank of harness variants under gated verification.
More recently, harnesses adapt during policy training~\citep{jiang2026harness1,zhang2026hase,chen2026evotrainer,yu2026coevolvingharness}. EvoTrainer~\citep{chen2026evotrainer} co-evolves the policy with a training harness that diagnoses and intervenes on recorded runs, and harness-and-model co-evolution~\citep{yu2026coevolvingharness} corrects on-policy rollouts as the harness changes.
\ourmethod{} is such a training-time harness, and its harness memory supplies the run's rollout and update history as evidence. It differs in that the JIT critic measures each primary candidate update on the current frozen policy before accepting it into $\harness{}_{t+1}$.

\section{Conclusion}

We introduce \ourmethod{}, a training-time harness that co-evolves with the policy through task sampling, rollout guidance, and rubric criteria. The harness memory provides within-task history and cross-task evidence, which the harness evolver uses to attribute each reward failure to the interface whose update can address it. The JIT critic then accepts a candidate update only after measuring its effect on the current frozen policy. We also curate \corpus{}, a training and evaluation suite of training tasks with seed rubrics across the medical, science, writing, role-play, and instruction-following domains, together with held-out tasks and an OOD benchmark for each domain. Extensive results show that \ourmethod{} outperforms the strongest baseline on both in-domain and OOD evaluations at both model scales. Its training dynamics show lower failure-class proportions and higher reward-oracle agreement than GRPO, with reversed rubric updates remaining rare, and its efficiency allows it to reach its best OOD checkpoint in fewer cluster hours than any baseline and in about half the steps GRPO needs.

\section*{Acknowledgement}
This work was supported by Alibaba Research Intern Program.

\bibliographystyle{colm2024_conference}
\bibliography{references}

\end{document}

%% file: commands.tex
\definecolor{dkgreen}{rgb}{0,0.6,0}
\definecolor{gray}{rgb}{0.5,0.5,0.5}
\definecolor{mauve}{rgb}{0.58,0,0.82}

\definecolor{dgreen}{rgb}{0.412,0.741,0.271}
\definecolor{dblue}{rgb}{0.220,0.325,0.639}
\definecolor{dred}{rgb}{0.933,0.122,0.137}

\definecolor{bestcell}{RGB}{232,245,233}    %
\definecolor{rowshade}{RGB}{248,248,248}    %
\definecolor{csframe}{RGB}{46,125,50}       %
\definecolor{csbg}{RGB}{250,253,250}        %
\definecolor{darkgreen}{RGB}{27,94,32}

\definecolor{g1}{HTML}{b3e2cd}
\definecolor{r1}{HTML}{fdcdac}
\definecolor{w1}{HTML}{cbd5e8}
\definecolor{b1}{HTML}{fff7bc}

\definecolor{lr}{HTML}{bebada}
\definecolor{fr}{HTML}{fccde5}

\definecolor{Lavender}{HTML}{BF94E4}

\newcommand{\ie}{\textit{i}.\textit{e}.\,}

\definecolor{l1}{RGB}{189,215,238}
\definecolor{l2}{RGB}{222,235,247}
\definecolor{l3}{RGB}{255,230,153}
\definecolor{l4}{RGB}{248,203,173}
\definecolor{l5}{RGB}{244,177,131}

\newtcbox{\columntcbox}{highlight math style={
        colback=gray!30,
        arc=2pt,
        outer arc=2pt,
        boxrule=0pt,
        top=2pt,
        bottom=2pt,
        left=2pt,
        right=2pt,
    }
}

\colorlet{LightLavender}{Lavender!35!}
\newtcbox{\inlinetcbox}[1][]{on line,
        boxsep=2pt, left=0pt,right=0pt,top=0pt,bottom=0pt,
        colframe=white,colback=LightLavender,
        highlight math style={enhanced}, #1
}

\newcolumntype{C}[1]{>{\centering}m{#1}}

\makeatletter
\newcommand{\xMapsto}[2][]{\ext@arrow 0599{\Mapstofill@}{#1}{#2}}
\def\Mapstofill@{\arrowfill@{\Mapstochar\Relbar}\Relbar\Rightarrow}
\makeatother

\definecolor{cSamp}{HTML}{2F5597}    %
\definecolor{cGuid}{HTML}{3C7A3C}    %
\definecolor{cRubr}{HTML}{B4622A}    %
\definecolor{cSat}{HTML}{A0803F}     %
\definecolor{cSpur}{HTML}{9E565A}    %
\definecolor{cBlind}{HTML}{5E7D95}   %

\newcommand{\opn}[2]{{\color{#1}\texttt{#2}}}
\newcommand{\osamp}[1]{\opn{cSamp}{#1}}
\newcommand{\oguid}[1]{\opn{cGuid}{#1}}
\newcommand{\orubr}[1]{\opn{cRubr}{#1}}
\newcommand{\clname}[1]{\text{\ttfamily\footnotesize #1}}
\newcommand{\clblind}{\clname{blind}}
\newcommand{\clsat}{\clname{saturated}}
\newcommand{\clspur}{\clname{spurious}}
\newcommand{\clagree}{\clname{agree}}
\newcommand{\role}[1]{\textsc{#1}}

\newcommand{\hctrl}[1]{\textcolor{blue}{\boldsymbol{#1}}}
\definecolor{cMem}{HTML}{1C6B5E}     %

\newtcbox{\pspan}[1][]{on line, nobeforeafter,
  boxsep=2.2pt, left=1pt, right=1pt, top=0pt, bottom=0pt,
  arc=2.2pt, boxrule=0pt, colframe=white, #1}

\newcommand{\ptag}[2]{\text{{\color{#1}\ttfamily\footnotesize<#2>}}}

\newcommand{\gtag}[1]{\ptag{cGuid}{#1}}

\newcommand{\mtag}[1]{\ptag{cMem}{#1}}

\RequirePackage{alltt}
\newtcolorbox{promptbox}[2][]{enhanced jigsaw, breakable, pad before break*=3.5mm, pad after break=1.5mm,
  skin first=enhanced, skin middle=enhanced, skin last=enhanced,
  colback=white, colframe=black, colbacktitle=black, coltitle=white, coltext=black,
  boxrule=0.6pt, titlerule=0pt, arc=0pt, outer arc=0pt, toprule at break=0.6pt, bottomrule at break=0.6pt,
  boxsep=0pt, left=7pt, right=7pt, top=5pt, bottom=5pt, toptitle=2.5pt, bottomtitle=2.5pt,
  before skip=8pt, after skip=3pt, middle=4pt,
  fonttitle=\small\bfseries, title={#2}, fontupper=\small,
  #1}

\newcounter{pstep}
\makeatletter
  {\endlist}
\makeatother


\newcommand{\psep}{\par\vskip4pt\cleaders\hbox to\linewidth{\hskip-7pt%
  \vrule width\dimexpr\linewidth+14pt\relax height0.4pt depth0pt\hss}\vskip0.4pt\vskip4pt\relax}